\documentclass[runningheads]{llncs}
\usepackage{graphicx}
\usepackage{comment}
\usepackage{amsmath,amssymb} 
\usepackage{color}

\usepackage{booktabs}
\usepackage{amsfonts}
\usepackage{mathrsfs}
\usepackage{multirow}
\usepackage{subfigure}
\usepackage{colortbl}
\usepackage{tabu}
\usepackage{bbm}

\begin{document}

\pagestyle{headings}
\mainmatter

\title{Prototype Refinement Network \\ for Few-Shot Segmentation}

\titlerunning{Prototype Refinement Network for Few-Shot Segmentation}
%
\author{Jinlu Liu \and
Yongqiang Qin}
\authorrunning{J. Liu and Y. Qin}
%
\institute{AInnovation Technology Co., Ltd. \\
\email{liujinlu, qinyongqiang@ainnovation.com}}

\maketitle

\begin{abstract}
Few-shot segmentation targets to segment new classes with few annotated images provided. It is more challenging than traditional semantic segmentation tasks that segment known classes with abundant annotated images. In this paper, we propose a Prototype Refinement Network (PRNet) to attack the challenge of few-shot segmentation. It firstly learns to bidirectionally extract prototypes from both support and query images of the known classes. Furthermore, to extract representative prototypes of the new classes, we use \textit{adaptation} and \textit{fusion} for prototype refinement. The step of adaptation makes the model to learn new concepts which is directly implemented by retraining. Prototype fusion is firstly proposed which fuses support prototypes with query prototypes, incorporating the knowledge from both sides. It is effective in prototype refinement without importing extra learnable parameters. In this way, the prototypes become more discriminative in low-data regimes. Experiments on PASAL-$5^i$ and COCO-$20^i$ demonstrate the superiority of our method. Especially on COCO-$20^i$, PRNet significantly outperforms existing methods by a large margin of 13.1\% in 1-shot setting.
\end{abstract}

\section{Introduction}
Significant breakthroughs have been made in semantic segmentation tasks by taking the benefits of strong feature learning ability of deep neural networks \cite{simonyan2015very,he2016deep}. The performance of these neural networks such as FCN \cite{long2015fully} and RefineNet \cite{lin2017refinenet} is severely dependent on the large-scale training data which requires a large number of pixel-level annotations. However, pixel-level annotations are difficult and expensive to obtain and furthermore, training with abundant data makes the model hard to generalize to new categories with only few labeled data. It arouses the interest of learning a model that can segment a new concept from few samples, which is so-called few-shot segmentation. In a $N$-way $K$-shot few-shot segmentation task, we target to segment the query images given $K$ annotated support images from each of the $N$ classes.

Prototype learning based methods are widely used in handling few-shot tasks like classification \cite{snell2017prototypical} and segmentation \cite{dong2018few,wang2019panet}. For few-shot segmentation, these methods treat it as a classification problem that classifies each pixel by matching it with the nearest prototype. Existing methods \cite{dong2018few,wang2019panet} put emphasis on learning representative prototypes  in the feature space which are closer to the pixels belonging to the same semantic class. However, in few-shot segmentation, prototypes directly extracted from the support images are usually biased to represent the semantic classes in the query images, due to the data scarcity and large intra-class variance. Therefore, we are encouraged to refine the prototypes be more representative for segmentation through pixel-wise matching.

\begin{figure*}
\centering
\includegraphics[height=2in, width=5in]{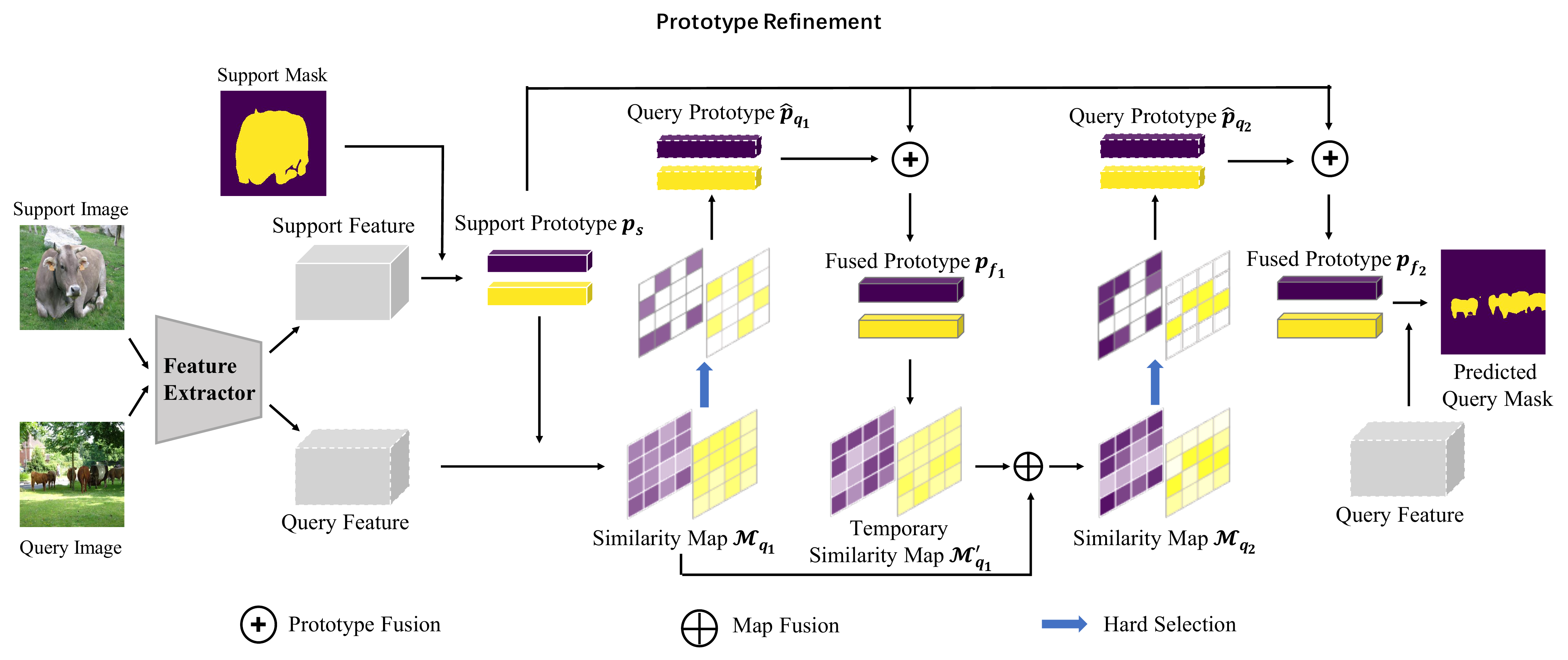}
\caption{We take a 1-way 1-shot task for example to illustrate our prototype refinement network for few-shot segmentation. \textit{Adaptation} and \textit{fusion} are used for prototype refinement. The feature extractor is adapted to the support set for prototype extraction. And we fuse the support prototypes with the predicted query prototypes as the final prototypes for segmentation. 
Color shades in the map indicate the similarity degree. Best viewed in color.
}
\label{figure:framework}
\end{figure*}

In this paper, we propose Prototype Refinement Network (PRNet) for few-shot segmentation. The proposed method mainly consists of two parts: prototype learning and prototype refinement. 
At the training stage, the network learns to extract prototypes from both support images and query images, which is different from most methods that merely extract prototypes from the support set. Images are input to the network embedded as the deep features and we obtain the support prototypes by masked average pooling on the support features. Then, the segmentation mask of the query image is predicted by nearest prototype matching of the query feature at each spatial location. Given the predicted mask, we also compute query prototypes to segment the support images in the same way. The bidirectional constraints drive the network to extract representative support prototypes that can precisely segment the query images and vice versa.

At the test stage, it requires the network to perform segmentation on the classes that are unseen during training. To extract more representative prototypes, we implement \textit{adaptation} and \textit{fusion} for prototype refinement as shown in Fig. \ref{figure:framework}. The feature extractor is retrained on the few support images to adapt to the unseen classes. For adaptation, the support prototypes are extracted to segment the support images which is different from the practice in training. Furthermore, we propose a two-stage fusion method that fuses the support prototypes with the query prototypes for refinement. We compute similarity maps where the confidently segmented regions are identified by a self-adaptive threshold $\alpha$. The query prototypes $\hat{p}_{q_1}$ are accordingly extracted from the spatial locations that are included in the selected regions. As displayed in Fig. \ref{figure:framework}, the fused prototype $p_{f_1}$ is firstly formed by support prototypes $p_s$ and query prototypes $\hat{p}_{q_1}$. Given $p_{f_1}$, we compute new similarity maps, obtaining new prototypes $\hat{p}_{q_2}$ after map fusion and hard selection. Then we implement fusion the second time by fusing $p_s$ with $\hat{p}_{q_2}$. In this way, the discriminative knowledge of the query images is introduced into prototype extraction. Prototype fusion is an efficient method for refinement without introducing extra learnable parameters. We conduct experiments on PASCAL-$5^i$ and two setups of COCO-$20^i$, finding that our proposed PRNet brings in an obvious improvement by large margins.

Our contributions are summarized as:
\begin{itemize}
    \item[1)] We propose a prototype refinement network for few-shot segmentation, which effectively explores the representative prototypes and perform segmentation by nearest prototype matching.
    \item[2)] We demonstrate the effectiveness and simplicity of the proposed fusion method that can significantly improve the performance without importing extra learnable parameters. The fused prototypes are more representative in few-shot scenarios.
    \item[3)] We achieve the state-of-the-art results on both PASCAL-$5^i$ and COCO-$20^i$. Especially on COCO-$20^i$, we consistently outperform existing methods in all cases with the increase of mean-IoU score up to 17.4\% in 5-shot setting.
\end{itemize}

\section{Related Works}
\paragraph{Semantic Segmentation} classifies each pixel in an image into several pre-defined classes. Abundant approaches are exploited for pixel-wise segmentation tasks such as FCN \cite{long2015fully}, UNet \cite{ronneberger2015u} and DeepLab series \cite{chen2018deeplab,chen2017rethinking,chen2018encoder}. Dilated convolution \cite{yu2016multi} is widely adopted in segmentation tasks to increase receptive fields which is also used in our methods.

\paragraph{Few-Shot Learning} aims to predict image labels with few support samples provided. Existing methods are commonly divided into two braches: meta-learning based methods \cite{finn2017model,nichol2018reptile} and metric learning based methods \cite{snell2017prototypical,allen2019infinite}. MAML \cite{finn2017model} is a typical meta-learning method which targets to learn a good weight initialization from small support samples, enabling fast adaptation to new tasks by a few gradient steps. PN \cite{snell2017prototypical} is a typical metric learning method which performs classification by finding the nearest class prototype in the embedding space.

\paragraph{Few-Shot Segmentation} is a new task that attacks the semantic segmentation problem in few-shot scenarios. It requires to perform pixel-wise segmentation from few annotated images available. Most existing approaches adopt two-brach architectures to align support images with query images \cite{shaban2017one,zhang2018sg,wang2019panet,hu2019attention}. OSLSM \cite{shaban2017one} was proposed for one-shot semantic segmentation which uses support branch to learn parameters for the logistic regression layer of the query branch. 
Prototype learning based methods \cite{dong2018few,wang2019panet} were recently proposed for few-shot segmentation. PANet \cite{wang2019panet} viewed the averaged features as prototypes and trained the feature extractor by adding alignment constraints between support prototypes and query prototypes. Our method is similar to PANet but we further propose the prototype refinement approach to make prototypes more suitable to represent the target classes in low-data regimes.

\section{Methods}
Our proposed method mainly consists of two stages: prototype learning and prototype refinement, which correspond to the training and test stages accordingly. The network firstly learns from the training classes to extract class prototypes from both support set and query set. When evaluating on the test classes, we tune the feature extractor to adapt to the support set and then, we use prototype fusion for further refinement. The prototypes refined in this way are more representative in few-shot scenarios.

\subsection{Problem Definition}
In few-shot segmentation, a dataset $\mathcal{D}_{train}$ of classes $\mathcal{C}_{train}$ is given for training the segmentation network. Then the network is required to perform segmentation on the set $\mathcal{D}_{test}$ of classes $\mathcal{C}_{test}$ which have no overlap with $\mathcal{C}_{train}$. Both training and test datasets are composed of \textit{N}-way \textit{K}-shot segmentation tasks $\mathcal{T}=\{ (\mathcal{S}, \mathcal{Q}) \}$.
In each task, the model learns from the support set $\mathcal{S}=\bigcup ^N\limits_{n=1} \{(I_s^{n,k}, M_s^{n,k})\} ^K_{k=1}$ where each of $N$ classes has $K$ support images $I_s$ paired with ground-truth segmentation masks $M_s$. And the model need to segment the images in the query set $\mathcal{Q}=\{(I_q^{t}, M_q^{t})\} ^T_{t=1}$ where $T$ is the total number of the query images in a task. At the training stage, we train our network on the tasks sampled from $\mathcal{D}_{train}$ while at the test stage, we evaluate it on the tasks from $\mathcal{D}_{test}$.

\subsection{Prototype Learning}
The general idea of existing prototypical networks \cite{snell2017prototypical,dong2018few,wang2019panet} is representing categories by the learnt feature vectors which are viewed as class prototypes in the feature space. According to it, pixel-wise segmentation can be performed by labeling every pixel with the semantic class of the nearest prototype. Inspired by \cite{wang2019panet}, we train the model by bidirectionally learning prototypes from both support images and query images. 
Support and query images are first input to the network $f_\theta$ embedded as the deep features. Based on the features, support prototypes of foreground classes are computed by masked average pooling $MAP(\cdot)$ which are denoted as:
\begin{equation}
\begin{split}
p^n_s & =\frac{1}{K} \sum ^K_{k=1} MAP(f_\theta (I_s^{n,k}),M_s^{n,k})  \\
& = \frac{1}{K} \sum^K_{k=1} \frac{\sum_{x,y} F^{n,k}_{s;x,y} \ \mathbbm{1}(M_{s;x,y}^{n,k} = n ) }{\sum_{x,y} \mathbbm{1}(M_{s;x,y}^{n,k} = n ) }
\end{split}
\end{equation}
where $F^{n,k}_{s}$ is the extracted support feature and $(x,y)$ is the spatial location in the feature.
Background is also treated as a semantic class whose prototype $p^0_s$ is computed by feature averaging of all spatial locations excluded in any foreground classes.
Given the support prototypes $\{p_s^n\}^{N}_{n=0}$ of $N$ foreground classes and the background, we perform segmentation by distance computation between the query features at each spatial location and the support prototypes. The segmentation mask $\hat{M}_{q}$ of the query image is predicted pixel by pixel as follows:
\begin{equation}
\begin{split}
\hat{M}^t_{q;x,y} & = \arg\max_n \; \mathcal{M}^{t,n}_{q;x,y} \\
& = \arg\max_n \; \sigma (d(F^t_{q;x,y}, p_s^n))
\end{split}
\label{eq:query_mask}
\end{equation}
where $F^t_{q;x,y}$ is the feature vector of query image $I^t_q$ at the location $(x,y)$ and  $\sigma(\cdot)$ is the softmax operation. $d(\cdot)$ is the cosine similarity function. If the query image is correctly segmented, there should be query prototypes extracted accordingly that can segment the support images as described above. Therefore, we extract query prototypes $\hat{p}_q$ based on the predicted mask $\hat{M}_q$ and perform segmentation on the support images as follows:
\begin{equation}
\hat{p}_q=\frac{1}{T} \sum ^T_{t=1} MAP(f_\theta (I_q^{t}),\hat{M}_q^{t})
\end{equation}
\begin{equation}
\hat{M}_s = \arg\max_n \; \sigma (d(F^k_{s}, p_q^n))
\end{equation}  
The network is trained end-to-end by the objective:
\begin{equation}
\mathcal{L}_{train}= \mathcal{L}_{CE}(M_s,\mathcal{M} _s) + \mathcal{L}_{CE}(M_q,\mathcal{M} _q)
\label{eq:train}
\end{equation} 
where $\mathcal{L}_{CE}(\cdot)$ is the standard cross-entropy loss. 
We regulate the network from both sides as indicated in Eq. \ref{eq:train}.
In this way, the network is driven to learn an efficient feature space where the prototypes are more representative to the semantic classes.

\subsection{Prototype Refinement}
\label{sec:refine}
Since the test classes $\mathcal{C}_{test}$ are unseen at the training stage, the prototypes directly extracted from few support images are inevitably to be biased. It is to say that the support prototypes are less capable of representing the semantic class due to the data scarcity and class variance. Therefore, we propose to refine class prototypes at the test stage by two strategies: adaptation and fusion.

\textbf{Adaptation} 
Since $\mathcal{C}_{test}$ remains unseen to the feature extractor $f_\theta$ before testing, it is necessary to adapt $f_\theta$ to $\mathcal{D}_{test}$ for discriminative feature extraction. Retraining a meta-trained model on a few-shot set is verified to achieve strongly competitive performance in low-data regimes \cite{dhillon2020a}. Referring to it, we retrain the feature extractor on the support set by the following objective:
\begin{equation}
\mathcal{L}_{adapt}= \mathcal{L}_{CE}(M_s,  \mathcal{M}_s)
\label{eq:adapt}
\end{equation} 
where $\mathcal{M}_s= \sigma(d(F_s, p_s))$. Different from the practice at the prototype learning stage, we use support prototypes to segment support images as indicated in Eq. \ref{eq:adapt}. Retrained on the support set, the network can extract more discriminative features of the new classes.

\textbf{Fusion}
Due to the data scarcity and large intra-class variance in few-shot scenarios, the support prototypes are not representative enough in the embedding space.
Thus, the pixels in query images are easily to be misclassified by non-parametric distance computation with support prototypes. To make the prototypes be more representative, we propose a two-step fusion method for further prototype refinement. The framework of prototype fusion is shown in the right part of Fig. \ref{figure:framework}.
Given the support prototypes $p_s$, we compute cosine similarity map $\mathcal{M}_{q_1}$ of each query image by dense comparison with $p_s$ at each spatial location and predict the query mask $\hat{M}_{q_1}$. 
Similar to the practice at the prototype learning stage, query prototypes can be extracted accordingly. However, the rough segmentation of query images will result in the ineffectiveness of the extracted query prototypes. Thus we apply hard selection $\mathcal{H}(\cdot)$ with a self-adaptive threshold $\alpha$ to segment high-confident regions, which is to say that, locations with cosine similarity larger than $\alpha$ are considered to be correctly segmented. The selected locations form the sparse query mask $ \mathcal{H}(\mathcal{M}_{q_1}) $ that is used for masked average pooling to extract query prototype $\hat{p}_{q_1}$ as follows: 
\begin{equation}
\hat{p}_{q_1}  =\frac{1}{T} \sum ^T_{t=1} MAP(f_\theta ' (I^t_q), \mathcal{H}(  \hat{M}^t_{q_1}  , \mathcal{M}^t_{q_1}) )
\end{equation}
where $f' _\theta$ is the adapted feature extractor. 
The hard selected is implemented by: 
\begin{equation}
\mathcal{H}( \hat{M}^t_{q_1}, \mathcal{M}^t_{q_1} ) _{x,y} =  \begin{cases}
\hat{M}^t_{q_1;x,y} & if \mathcal{M}^{t,n}_{q_1;x,y} > \alpha ^n  \\
-1 & otherwise
\end{cases}
\end{equation}
where the value $-1$ indicates the location not belonging to the confident regions and $n={\arg\max}_j \mathcal{M}^{t,j}_{q_1;x,y}$ where $j=0,...,N$. The value of threshold $\alpha ^n$ of each class is given in Section \ref{section:experiment}.
Then we fuse support prototype $p_s$ with query prototype $\hat{p}_{q_1}$ by Eq. \ref{eq:fuse1}:
\begin{equation}
p_{f_1} = \omega _s p_s + \omega _q \hat{p}_{q_1}
\label{eq:fuse1}
\end{equation}
where $\omega _s$ and $\omega _q$ are hyper-parameters that control the fusion weights. After obtaining the fused prototypes $p_{f_1}$, we compute a temporary similarity map $\mathcal{M}'_{q_1}$ through $p_{f_1}$, and apply map fusion on $\mathcal{M}_{q_1}$ and $\mathcal{M}'_{q_1}$ to get the final map: 
\begin{equation}
\mathcal{M}_{q_2} = (\mathcal{M}_{q_1}+ \mathcal{M}'_{q_1}) / 2
\end{equation}
Then we obtain $\hat{M}_{q_2}$ as aforementioned. The query prototypes $\hat{p}_{q_2}$ are computed as:
\begin{equation}
\hat{p}_{q_2}  =\frac{1}{T} \sum ^T_{t=1} MAP(f_\theta ' (I^t_q), \mathcal{H}( \hat{M}^t_{q_2} ,  \mathcal{M}^t_{q_2})) 
\end{equation}
The second step of prototype fusion is conducted by:
\begin{equation}
p_{f_2} = \omega _s p_s + \omega _q \hat{p}_{q_2}
\end{equation}
Based on the refinement implemented so far, we utilize the fused prototype $p_{f_2}$ as the final prototype to segment the query image.

It can be seen that the prototypes $p_{f_2}$ fuse the knowledge from both support prototypes and query prototypes, which enjoy the advantage of both sides. On the one hand, remaining the knowledge of support images avoids overfitting when segmenting the query images. On the other hand, importing the knowledge from query images makes the prototype be more discriminative and representative.


\section{Experiments}
\label{section:experiment}
\subsection{Setup}
\textbf{Datasets}
We evaluate our model on PASCAL-$5^i$ \cite{shaban2017one} and COCO-$20^i$ \cite{nguyen2019feature}. PASCAL-$5^i$ is a derivation of PASCAL VOC 2012 \cite{everingham2010the} which is split into 4 folds. Each fold contains 5 categories. COCO-$20^i$ is a larger dataset which is derived from MS COCO \cite{lin2014microsoft}. 80 categories are divided into 4 folds with each fold containing 20 categories. Two splits are adopted in our experiments which are split-A \cite{nguyen2019feature} and split-B \cite{hu2019attention} respectively. All images are resized to 417x417.

\textbf{Evaluation Protocols}
We use mean-IoU and binary-IoU as evaluation protocols. Mean-IoU measures the average IoU score of all foreground classes. Binary-IoU treats all foreground objects as one class and the background is viewed as one class.

\textbf{Implementation Details}
Models are trained on three folds and evaluated on the rest fold. We use ResNet 101 \cite{he2016deep} as feature extractor. Following the previous works \cite{shaban2017one,wang2019panet}, the base network is initialized by the pretrained weights on ILSVRC \cite{russakovsky2015imagenet}. The network is trained end-to-end by SGD. At the test stage, most previous methods randomly sample 1,000 episodes for evaluation but the results vary greatly with different random seeds. To give a reliable and stable result, we follow \cite{wang2019panet} to report the average result of 5 runs. The number of query image is 1 as in \cite{shaban2017one,wang2019panet,nguyen2019feature}.

\textbf{Threshold}
The threshold $\alpha$ of each class is self-adaptively computed by:
$\alpha^n = (\mathcal{M}^n_{max}+\mathcal{M}^n_{mean})/2$
where 
$\mathcal{M}^n_{max} = max\{\mathcal{M}^n_{x,y}\}$ and $\mathcal{M}^n_{mean} = mean \{\mathcal{M}^n_{x,y}\}$.


\begin{table*}[h]
\centering
\resizebox{\textwidth}{21mm}{
\begin{tabular}{ccccccccccc}
\hline
 \multirow{2}{*}{\textbf{Methods}} &  \multicolumn{5}{c}{\textbf{1-shot}}  & \multicolumn{5}{c}{\textbf{5-shot}} \\ 
& Fold-1 & Fold-2 & Fold-3 & Fold-4 & \textbf{Mean} &  Fold-1 & Fold-2 & Fold-3 & Fold-4 & \textbf{Mean} \\ 
\hline
\hline
OSLSM \cite{shaban2017one} &  33.60 & 55.30 & 44.90 & 33.50 & 40.80 & 35.90 & 58.10 & 42.70 & 39.10 & 43.90 \\
co-FCN \cite{rakelly2018conditional} &  36.70 & 50.60 & 44.90 & 32.40 & 41.10 & 37.50 & 50.00 & 44.10 & 33.90 & 41.40 \\
SG-One \cite{zhang2018sg}  & 40.20 & 58.40 & 48.40 & 38.40 & 46.30 & 41.90 & 58.60 & 48.60 & 39.40 & 47.10 \\
PANet \cite{wang2019panet} & 42.30 & 58.00 & 51.10 & 41.20 & 48.10 & 51.80 & 64.60 & 59.80 & 46.50 & 55.70 \\
AMP \cite{siam2019amp}& 41.90 & 50.20 & 46.70 & 34.70 & 43.40 & 41.80 & 55.50 & 50.30 & 39.90 & 46.90  \\
CANet \cite{zhang2019canet}& 52.50 & 65.90 & 51.30 & 51.90 & 55.40 & 55.50 & 67.80 & 51.90 & 53.20 & 57.10 \\
FWB \cite{nguyen2019feature} &  51.30 & 64.69 & 56.71 & 52.24 & \textbf{56.19} & 54.84 & 67.38 & 62.16 & 55.30 & 59.92 \\ 
PRNet(ours)  & 51.60 & 61.32 & 53.10 & 47.59 & 53.40 & 57.84 & 67.31 & 64.08 & 53.36 & \textbf{60.56} \\ 
\hline
\end{tabular}}
\caption{Mean-IoU results on PASCAL-$5^i$.} 
\label{table:pascal-1way-miou}
\end{table*}

\begin{table}[h]
\centering

\begin{tabular}{cccccccccc}
\hline
 & OSLSM  & co-FCN  &  PL & A-MCG & SG-One & AMP & PANet  & CANet   &PRNet (ours)  \\  \hline
1-shot & 61.3 & 60.1 & 61.2 &  61.2 & 63.9 & 62.2 & 66.5 & 66.2 & \textbf{68.79}  \\
5-shot & 61.5 & 60.2 & 62.3 & 62.2 & 65.9 & 63.8 & 70.7 & 69.6 & \textbf{73.40} \\
 \hline
\end{tabular}
\caption{Binary-IoU results on 1-way PASCAL-$5^i$.} 
\label{table:pascal-1way-biou}
\end{table}

\subsection{1-way Segmentation}
\textbf{PASCAL-$5^i$} Table \ref{table:pascal-1way-miou} shows the mean-IoU results on PASCAL-$5^i$. 
Our PRNet achieves the best performance in the 5-shot scenarios. However, in the 1-shot setting, our result is relatively lower than CANet and FWB. One possible reason is that the adaptation step plays a limited role in the 1-shot tasks, resulting in small improvements. 
The binary-IoU scores on PASCAL-$5^i$ are displayed in Table \ref{table:pascal-1way-biou}. We can see that our method shows the best performance evaluated by binary-IoU.

\textbf{COCO-$20^i$} Table \ref{table:coco-1way-miou} compares our method with PANet and FWB on COCO-$20^i$ in different splits. 
It is obvious to see that our PRNet consistently outperforms the state-of-the-art methods on all folds. COCO-$20^i$ is a more challenging dataset that contains more categories and the quality of segmentation mask is relatively lower. However, our PRNet achieves significant improvements by a large margin especially on split-B. The detailed results on COCO derivatives are shown in Table \ref{table:coco-1way-biou}.

\begin{table*}[h]
\centering
\resizebox{\textwidth}{13mm}{
\begin{tabular}{cccccccccccc}
\hline
\multirow{2}{*}{\textbf{Methods}} & \multirow{2}{*}{\textbf{Split}}  & \multicolumn{5}{c}{\textbf{1-shot}}  & \multicolumn{5}{c}{\textbf{5-shot}} \\ 
& & Fold-1 & Fold-2 & Fold-3 & Fold-4 & \textbf{Mean} &  Fold-1 & Fold-2 & Fold-3 & Fold-4 & \textbf{Mean} \\ \hline
\hline
PANet \cite{wang2019panet} & \multirow{2}{*}{A}  & - & - & - & - & 20.90 & - & - & - & - & 29.70 \\
PRNet(ours) &  &41.16 & 26.28 &24.74 &21.92 & \textbf{28.53}  & 50.37 & 32.85 & 31.77 & 29.14 & \textbf{36.03} \\  
\hline 
FWB \cite{nguyen2019feature} &  \multirow{2}{*}{B}   &16.98 & 17.98 & 20.96 & 28.85 & 21.19 & 19.13 & 21.46 & 23.93 & 30.08 & 23.65 \\ 
PRNet(ours)  & &33.59 & 37.39 & 33.64 & 32.53 & \textbf{34.29} & 40.44 & 44.43 & 39.41 & 39.91 & \textbf{41.05} \\ 
\hline
\end{tabular}}
\caption{Mean-IoU results on COCO-$20^i$. Split-A and split-B are two different setups for 4-fold cross-validation on COCO-$20^i$.} 
\label{table:coco-1way-miou}
\end{table*}

\begin{table*}[h]
\centering
\begin{tabular}{cccc}
\hline
Methods & Split & 1-shot & 5-shot \\  \hline
A-MCG & \multirow{3}{*}{A} & 52.0 & 54.7 \\
PANet & & 59.2 & 63.5 \\
PRNet (ours) & & \textbf{61.40} & \textbf{64.69} \\
\hline
PRNet (ours) & B & \textbf{65.91} & \textbf{69.37} \\
\hline
\end{tabular}
\caption{Binary-IoU results on 1-way COCO-$20^i$.} 
\label{table:coco-1way-biou}
\end{table*}

\begin{figure}[h]
\centering
\includegraphics[height=3.5in, width=4in]{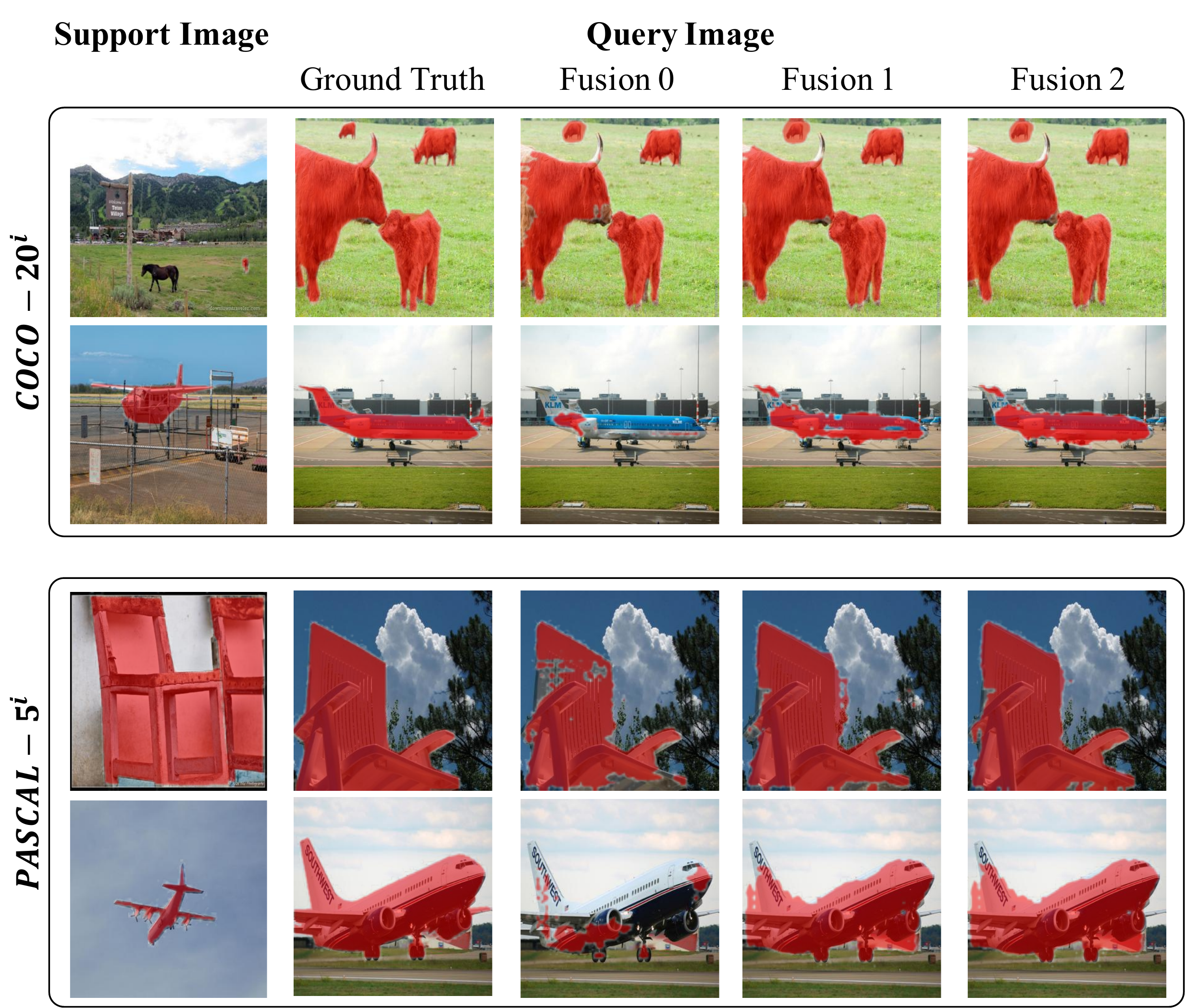}
\caption{1-shot qualitative results of prototype fusion. Best viewed in color with zoom.}
\label{figure:mask}
\end{figure}

\subsection{Qualitative Results}
The visualization of 1-shot segmentation is shown in Fig. \ref{figure:mask}. It can be seen that few-shot segmentation by prototype learning without refinement shows unsatisfactory results, since the prototypes extracted from one support image have limited generality to represent the semantic classes. As expected, our proposed prototype fusion is capable of segmenting more precise masks as shown in the figure. The first fusion step can segment large positive region of the target class which is wrongly segmented as background before. It is obviously displayed in the \textit{airplane} pictures. 

\subsection{Ablation Study}
\textbf{Backbone} We compare the results on COCO-$20^i$-B with different backbones in Table \ref{table:coco-backbone}. The ResNet series show better performance on few-shot segmentation tasks than VGG. It can be seen that even though we use VGG 16 as our backbone, PRNet achieves higher results than FWB which uses ResNet 101 in their experiments. The results on PASCAL-$5^i$ and COCO-$20^i$-A are given in the appendix.
\begin{table*}[h]
\centering

\begin{tabular}{ccccccccccc}
\hline
\multirow{2}{*}{\textbf{Backbone}}  & \multicolumn{5}{c}{\textbf{1-shot}}  & \multicolumn{5}{c}{\textbf{5-shot}} \\ 
&  Fold-1 & Fold-2 & Fold-3 & Fold-4 & \textbf{Mean} &  Fold-1 & Fold-2 & Fold-3 & Fold-4 & \textbf{Mean}  \\ \hline
VGG 16 & 27.46 & 32.99 & 26.70 & 28.98 & 29.03 & 31.18 & 36.54 & 31.54 & 32.00 & 32.82 \\
ResNet 50  & 31.24 & 35.49 & 29.98 & 31.93 & 32.16 & 32.06 & 38.3 & 32.65 & 33.22 & 34.06\\
ResNet 101 & 33.17 & 37.47 & 33.40 & 32.64 & 34.17 & 39.56 & 43.27 & 39.46 & 38.74 & 40.26 \\ 
\hline
\end{tabular}
\caption{Mean-IoU results on COCO-$20^i$-B with different backbones. The average results of 5 runs are reported without adaptation.} 
\label{table:coco-backbone}
\end{table*}

\begin{table}[h]
\centering
\begin{tabular}{cccc}
\hline
Adaptation  & Fusion  &  Mean-IoU & Binary-IoU \\  \hline
  &  & 48.62 & 71.36 \\
 \checkmark & & 50.17  & 72.76 \\
  & \checkmark & 49.25  & 71.18 \\
 \checkmark & \checkmark & 51.15 & 72.88 \\
 \hline
\end{tabular}
\caption{Ablation results on 1-way 5-shot COCO-$20^i$-A. We report the 1-run results of 1,000 randomly sampled episodes on Fold-1.} 
\label{table:refinement-ablation}
\end{table}

\begin{figure*}[h]
\centering
\subfigure[Adaptation]{
\begin{minipage}[t]{0.5\linewidth}
\centering
\includegraphics[width=2.5in, height=1.8in]{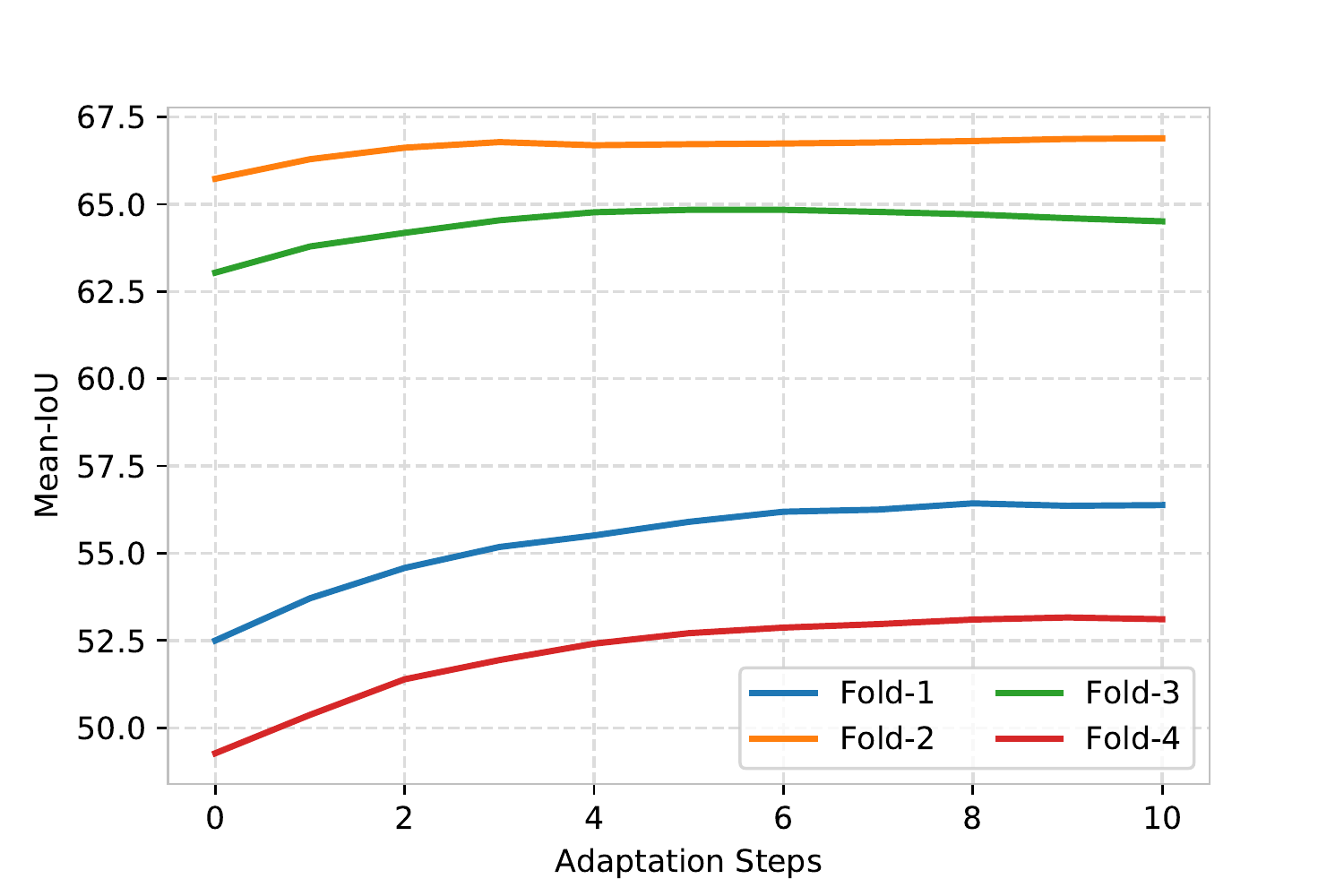}
\label{figure:refinement_steps}
\end{minipage}
}%
\subfigure[Fusion]{
\begin{minipage}[t]{0.5\linewidth}
\centering
\includegraphics[width=2.5in, height=1.8in]{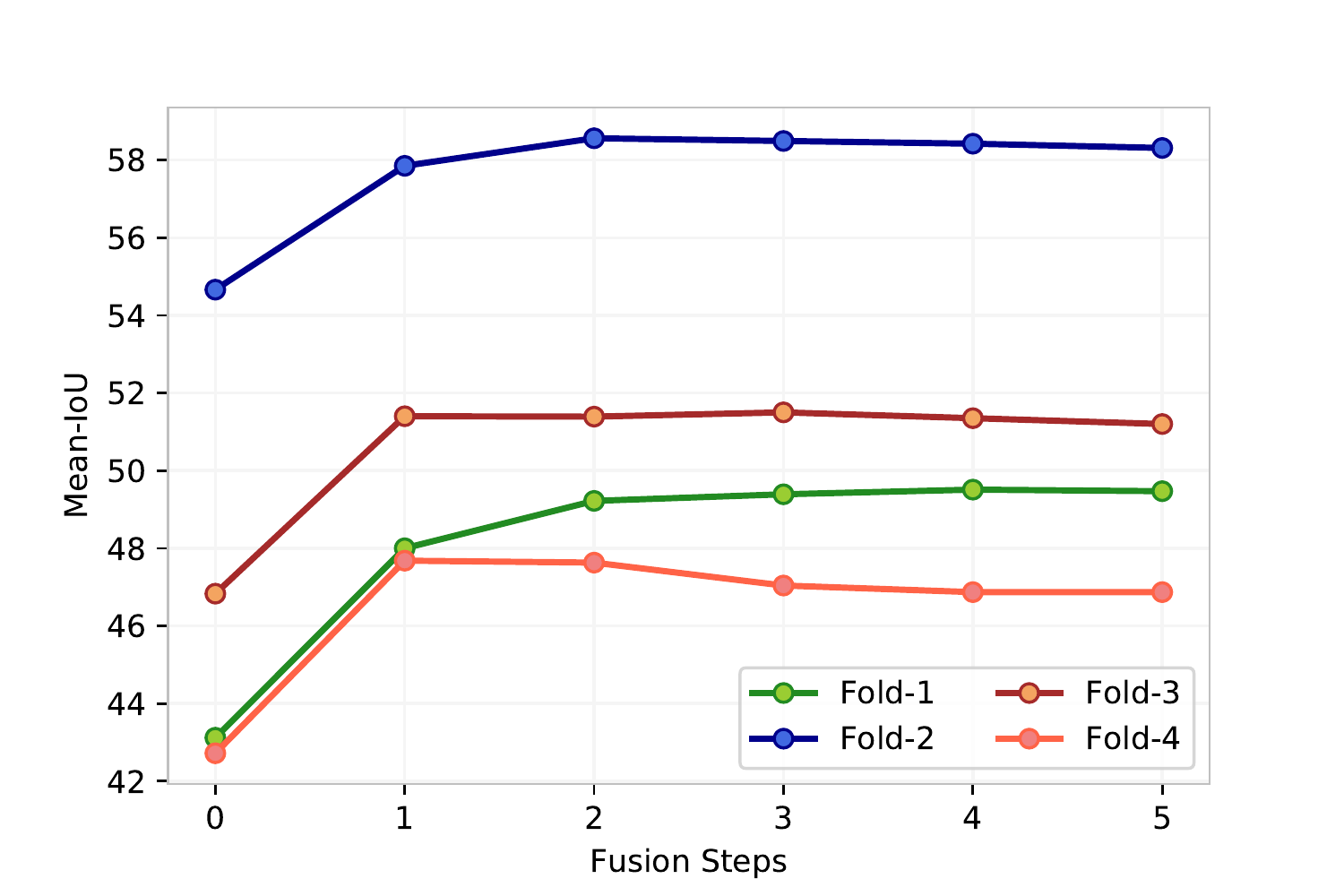}
\label{figure:fusion}
\end{minipage}%
}%
\caption{Ablation results of 1 random run on (a) 5-shot and (b) 1-shot PASCAL-$5^i$.}
\label{}
\end{figure*}

\textbf{Adaptation}
The results of adaptation are shown in Fig. \ref{figure:refinement_steps} which are computed from 1 run of 1,000 randomly sampled episodes in each fold. The adaptation yields consistent improvement with increasing steps and the overall performance reaches the summit after 5 steps. 
In our experiments, we take 5 adaptation steps for prototype refinement on the test fold.

\textbf{Fusion} 
Theoretically, the prototype fusion can be repeated more times. To illustrate the effectiveness of the proposed prototype fusion, we display the 1-run results in Fig. \ref{figure:fusion}. It can be seen that fusion after the second step brings in limited improvements and trends to result in a performance drop.
In practice, we implement the two-step fusion as explained in Section \ref{sec:refine}, which not only avoids overfitting but also imports more discriminative features into prototype computation.
We further give the ablation results on COCO-$20^i$ in Table \ref{table:refinement-ablation}.

\begin{table*}
\centering

\begin{tabular}{cccccccccccc}
\hline
$\omega_s$ & 0 & 0.1 & 0.2 & 0.3 & 0.4 & 0.5 & 0.6 & 0.7 & 0.8 & 0.9 & 1 \\
$\omega_q$ & 1 & 0.9 & 0.8 & 0.7 & 0.6 & 0.5 & 0.4 & 0.3 & 0.2 & 0.1 & 0 \\ \hline
Mean-IoU & 57.63 & 57.76 & 57.87 & 57.95 & 57.93 & 57.94 & 57.79  & 57.52 & 57.07 & 56.42 & 55.42\\
Binary-IoU & 71.53 & 71.52 & 71.57 & 71.66 & 71.79 & 71.93 & 72.08 & 72.23 & 72.37 & 72.48 & 72.45 \\
\hline 
\end{tabular}
\caption{Results of the one-step fusion on 1-shot PASCAL-$5^i$ with different $\omega_s$ and $\omega_q$.}
\label{table:pascal-weight}
\end{table*}

\textbf{Weights}
Table \ref{table:pascal-weight} compares the results with different weights $\omega_s$ and $\omega_q$. We report 1-run results on Fold-2, using VGG 16 as our backbone. The fusion weights are all set to 0.5 in this paper for the trade-off between the mean-IoU results and the binary-IoU results.

\section{Conclusions}
We propose the Prototype Refinement Network (PRNet) for few-shot segmentation. PRNet refines class prototypes by adaptation and fusion which makes the prototypes to be more representative. Furthermore, by simple fusion of support prototypes and query prototypes, we find it effective to bring in significant improvements for segmentation without importing extra learnable parameters. 
Comprehensive experiments show the superiority of our PRNet that consistently outperforms the state-of-the-art methods on COCO-$20^i$ by large margins. 

\appendix

\section{Implementation Details}
The network is trained end-to-end by SGD with learning rate of 7e-3, momentum of 0.9 and weight decay of 5e-4. The learning rate is reduced by 0.1 every 10,000 iterations. We train 30,000 iterations on PASCAL-$5^i$ and 20,000 iterations on COCO-$20^i$. Horizontal flipping is used for data augmentation. All experiments are implemented on PyTorch library. For adapting the models on the test fold, we use Adam optimizer with learning rate of 1e-6.

We modify the ResNet network as follows. The first two blocks remain unchanged. The strides of last two blocks are set to 1. To enlarge the reception field, we use dilated convolutions with rates of 2 and 4 in the last two blocks. Layers after the 4-th block are removed and, the last ReLU layer in the last block is removed.

\section{Backbone}
\begin{table}[h]
\centering

\begin{tabular}{ccccccccccc}
\hline
\multirow{2}{*}{\textbf{PASCAL-$5^i$}}  & \multicolumn{5}{c}{\textbf{1-shot}}  & \multicolumn{5}{c}{\textbf{5-shot}} \\ 
&  Fold-1 & Fold-2 & Fold-3 & Fold-4 & \textbf{Mean} &  Fold-1 & Fold-2 & Fold-3 & Fold-4 & \textbf{Mean} \\ \hline
VGG 16&  47.54 & 61.03 & 53.93 & 45.66 &  52.04 & 51.16 & 64.58 & 59.60 & 48.64 & 56.00 \\
ResNet 50 & 47.90 & 61.16 & 51.14 & 48.49 & 52.17 & 54.68 & 67.01 & 61.58 & 52.44 & 58.93 \\
ResNet 101 & 51.87 & 61.09 & 50.60 & 48.25 & 52.95 & 53.69 & 66.34 & 62.54 & 49.66 & 58.06 \\  
\hline
\end{tabular}
\caption{Mean-IoU results on PASCAL-$5^i$.} 
\label{table:pascal-1way-miou}
\end{table}

\begin{table}[h]
\centering
\newcommand{\tabincell}[2]{\begin{tabular}{@{}#1@{}}#2\end{tabular}}
\begin{tabular}{ccccccccccc}
\hline
 \multirow{2}{*}{\tabincell{c}{\textbf{COCO-$20^i$}\\ \textbf{split-A}}}  & \multicolumn{5}{c}{\textbf{1-shot}}  & \multicolumn{5}{c}{\textbf{5-shot}} \\ 
& Fold-1 & Fold-2 & Fold-3 & Fold-4 & \textbf{Mean} &  Fold-1 & Fold-2 & Fold-3 & Fold-4 & \textbf{Mean} \\ \hline
VGG 16 &  38.46 & 25.00 & 24.88 & 20.02  & 27.09  & 42.33 & 26.71 & 30.53 & 24.32 & 30.97 \\
ResNet 50 &  42.92 & 28.30 & 26.72 & 23.35 & 30.32 & 43.84 & 27.44 & 31.42 & 25.05 & 31.94  \\
ResNet 101 &  41.36	 & 24.86 & 25.23 & 21.11 & 28.14 & 46.95 & 28.83 & 31.77 & 25.93 & 33.37 \\  
\hline 
\end{tabular}
\caption{Mean-IoU results on COCO-$20^i$-A. } 
\label{table:coco-1way-miou-A}
\end{table}
To show the performance of PRNet with different backbones, we give the results on PASCAL-$5^i$ and COCO-$20^i$-A in Table \ref{table:pascal-1way-miou} and Table \ref{table:coco-1way-miou-A} respectively. Note that the results are reported without adaptation.

\section{2-way Segmentation}
\begin{table*}
\centering
\begin{tabular}{ccccccccccc}
\hline
\multirow{2}{*}{\textbf{Methods}} & \multicolumn{5}{c}{\textbf{1-shot}}  & \multicolumn{5}{c}{\textbf{5-shot}} \\ 
& Fold-1 & Fold-2 & Fold-3 & Fold-4 & \textbf{Mean} &  Fold-1 & Fold-2 & Fold-3 & Fold-4 & \textbf{Mean} \\ \hline
SG-One & - & - & - & - & - & - & - & - & - & 29.4\\
PANet & - & - & - & - & 45.1 & - & - & - & - & 53.1 \\
PRNet$_{fusion}$ & 43.49 & 55.24 & 48.32 & 39.92 & \textbf{46.74} &52.64&60.67&56.9&48.16& \textbf{54.59}\\  
\hline 
\end{tabular}
\caption{2-way Segmentation: mean-IoU score on PASCAL-$5^i$. In this experiment, we only use fusion for refinement without adaptation.} 
\label{table:pascal-2way-miou}
\end{table*}

Most methods just evaluate in 1-way tasks however, it should be extended to the tasks of more semantic classes. Without losing generality, we extend experiments on 2-way tasks on PASCAL-$5^i$. We only report the results of prototype fusion in Table \ref{table:pascal-2way-miou} for simply  illustrating the effectiveness of our method. In each 2-way task, a query image contains at least 1 foreground class. 
Despite the difficulty in 2-way tasks, we yield the best performance under all evaluation criteria. 
The results indicate the good generality of our method in higher ways setting. It can be easily extended to N-way K-shot tasks with retaining a stable performance. Notably, our 2-way mean-IoU scores are even higher than the results that some methods achieve in 1-way settings.

\bibliographystyle{splncs04}
\bibliography{PRNet}
\end{document}